\documentclass{egpubl}
\usepackage{pg2022}

\SpecialIssuePaper         

\usepackage[T1]{fontenc}
\usepackage{dfadobe}  

\usepackage{cite}  
\BibtexOrBiblatex
\electronicVersion
\PrintedOrElectronic
\ifpdf \usepackage[pdftex]{graphicx} \pdfcompresslevel=9
\else \usepackage[dvips]{graphicx} \fi

\usepackage{egweblnk} 

\usepackage{booktabs}  
\usepackage{mathptmx}
\usepackage{subcaption}
\usepackage{amsmath}
\usepackage{amssymb}
\title[BareSkinNet: De-makeup and De-lighting via 3D Face Reconstruction]%
      {BareSkinNet: De-makeup and De-lighting \\via 3D Face Reconstruction}

\author[Yang et al.]
{\parbox{\textwidth}{
 \centering Xingchao Yang\orcid{0000-0003-4736-1666}
         and Takafumi Taketomi\orcid{0000-0002-5353-0895} 
         }
         \\
{\parbox{\textwidth}{
\centering CyberAgent, AI Lab, Japan
}
}
}

\volume{41}   
\issue{7}     
\pStartPage{1}      

\begin{document}

\teaser{
 \includegraphics[width=\linewidth]{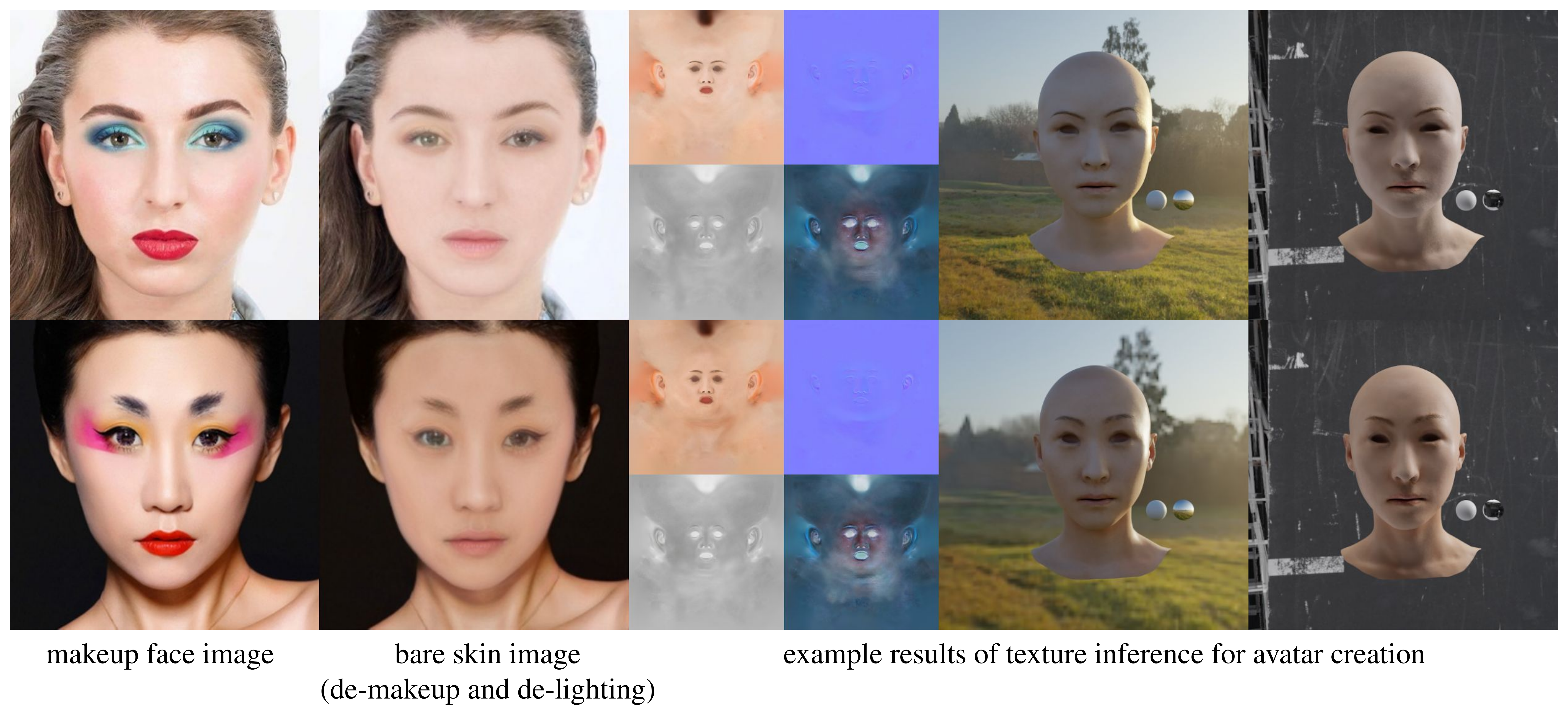}
 \centering
  \caption{Our method successfully removes makeup and lighting influences from the input face image to recover a bare skin (de-makeup and de-lighting) face image. The bare skin image facilitates subsequent applications such as the normalized texture (diffuse, normal, roughness, and specular) inference and avatar creation.}
\label{fig:teaser}
}

\maketitle
\begin{abstract}
   We propose BareSkinNet, a novel method that simultaneously removes makeup and lighting influences  from the face image. Our method leverages a 3D morphable model and does not require a reference clean face image or a specified light condition. By combining the process of 3D face reconstruction, we can easily obtain 3D geometry and coarse 3D textures. Using this information, we can infer normalized 3D face texture maps (diffuse, normal, roughness, and specular) by an image-translation network. Consequently, reconstructed 3D face textures without undesirable information will significantly benefit subsequent processes, such as re-lighting or re-makeup. In experiments, we show that BareSkinNet outperforms state-of-the-art makeup removal methods. In addition, our method is remarkably helpful in removing makeup to generate consistent high-fidelity texture maps, which makes it extendable to many realistic face generation applications. It can also automatically build graphic assets of face makeup images before and after with corresponding 3D data. This will assist artists in accelerating their work, such as 3D makeup avatar creation.


\end{abstract}  

\section{Introduction}

The realistic 3D avatar generation has become increasingly popular because of its ever-expanding applications in technologies such as virtual and augmented reality, video conferences, and games. With research efforts, a high-quality 3D face model can be acquired through consumer cameras or smartphones~\cite{Yamaguchi:2018:ToG, wu2020unsup3d, Bao:2021:ToG} and time-consuming specialized hardware~\cite{Debevec:2000:SIGGRAPH, Sun:2020:SIGGRAPHAsia, 10.1145/3386569.3392464}. 3D face reconstruction from portraits dramatically improves the convenience and speed of the avatar creation.

In general, high-fidelity avatar creation from portraits is achieved using deep-learning-based generative models~\cite{Gecer_2019_CVPR,Lattas_2020_CVPR,Yamaguchi:2018:ToG}. In these methods, a large-scale high-resolution texture dataset is used to train the networks through supervised learning. However, portrait-based high-fidelity 3D face reconstruction is still difficult when the face image includes complex environment lighting or large expressions. Many face normalization methods have been introduced to overcome these issues ~\cite{Nagano:2019:ToG, Luo_2021_CVPR}. The input face image is normalized to become as close as possible to be consistent with the high-resolution texture dataset. Although these methods show impressive results on environmental issues, such as pose, lighting, and expression, they do not consider artificial factors such as facial makeup. However, makeup is prevalent in daily life photos. It is necessary to be aware of how to normalize the face image with various kinds of makeup, from light to heavy.
Although enlarging the dataset will increase the ability to handle makeup~\cite{Scherbaum:2011:Makeup}, it is still difficult to address all of them.
Additionally, collecting numerous makeup faces is not a realistic task in controlled environments, such as the light stage.

\begin{figure*}[!ht]
    \centering
    \includegraphics[width=\textwidth]{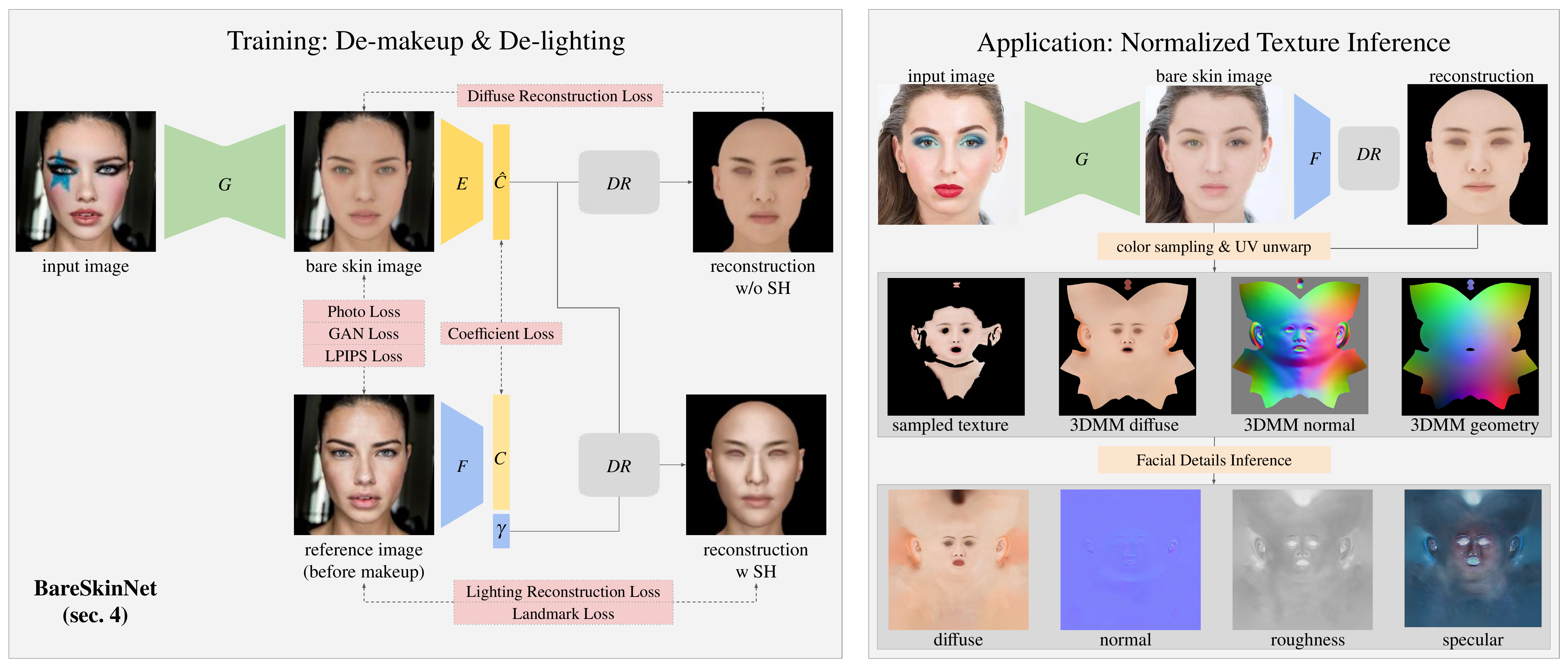}
    \caption{Overview of the proposed system. BareSkinNet comprises a de-makeup and de-lighting network $G$, a 3D face reconstruction network $E$, and a Pre-trained 3D face reconstruction network $F$. Makeup and lighting influences are removed from the input makeup image. In the inference stage, given an input makeup image, BareSkinNet outputs the bare skin image and the 3DMM coefficients. The diffuse, normal, roughness, and specular maps are inferred from the unwarped UVs of the bare skin image and reconstructed 3DMM.} 
\label{fig:01_method}
\end{figure*}

The study aims to generate face images without makeup and lighting influences as shown in Fig. \ref{fig:teaser}. Fig. \ref{fig:01_method} shows an overview and application of the proposed method. We introduce a de-makeup and de-lighting method that can generate bare skin images by utilizing 3D face reconstruction. Therefore, the corresponding 3D information can be easily accessed. As a result, we can obtain clean 3D face textures without makeup and lighting. The application of high-fidelity texture inference becomes more accurate and accessible.

For the makeup face image input, we propose to build a specialized network called BareSkinNet, which can estimate two types of information. 1) A  bare skin image is generated to preserve the high-frequency appearance information of the input face image. The bare skin color is consistent with the high-resolution texture dataset. 2) The low-frequency information is estimated by a 3D face reconstruction process of the 3D morphable model (3DMM).

To remove makeup influences from the input face image, the LADN dataset~\cite{gu2019ladn} is used to train BareSkinNet via weakly supervised learning. The lighting influences are removed by the process of 3D face reconstruction using 3DMM. Our experiments proved that by combining the process of 3D face reconstruction, the effectiveness of makeup removal is also enhanced. We further advanced the capabilities of BareSkinNet in a teacher-student manner. We can employ joint learning of BareSkinNet using a differentiable rendering technique.

In addition, we show an application of high-fidelity texture inference using the results of BareSkinNet. We use the 180 scanned face dataset captured by the high-quality multi-camera scan system to train the high-fidelity texture inference network. Scanned data include facial geometries and the corresponding 4K-resolution diffuse, normal, roughness, and specular texture maps. By combining BareSkinNet and the high-fidelity texture inference network, we can obtain a makeup- and lighting-free 3D face model.

Our contributions are summarized as follows:
\begin{itemize}
\item To the best of our knowledge, we are the first to explore how to remove makeup for face normalization. Our method removes both makeup and lighting influences to produce a bare skin image consistent with the texture dataset, which helps the subsequent process of  high-fidelity texture inference.
\item We propose BareSkinNet, the method jointly training a generator and a process of 3D face reconstruction by leveraging a 3D morphable model. This will be a great convenience because we do not require a reference clean face image or specified light conditions, which solves the problem of non-existent ground truth for in-the-wild de-makeup and de-lighting face images. A teacher-student manner of 3D face reconstruction is introduced to improve performance. We conducted a detailed ablation study to confirm the effectiveness of each component and loss function designs.
\item We demonstrate that BareSkinNet can stably produce bare skin images under various makeup and lighting conditions. We show the result of normalized texture inference for 3D avatar creation. The inferred texture maps can be used to create clean avatars for re-lighting and re-makeup processes.
\end{itemize}

\section{Related Work}
This study aims to generate a normalized 3D face model from a single image input under various makeup and lighting conditions. Therefore, in this section, we briefly review the existing works on 3D face reconstruction from single image input and face normalization techniques for constructing facial avatars.

\subsection{Image-based 3D Face Reconstruction}
3DMM has been widely used to reconstruct 3D face models~\cite{1227983, tran2016regressing, tewari2017mofa, GuoTPAMI2018, genova2018unsupervised, deng2020accurate, Bao:2021:ToG, 10.1145/3450626.3459936, EMOCA:CVPR:2022}. The 3D face model is reconstructed by fitting the 3DMM to the 2D face image input using facial features, such as face landmarks. Face 3DMM was first proposed by Blanz and Vetter~\cite{10.1145/311535.311556}. In general, 3DMM is created from a scanned face dataset~\cite{5279762, FaceWarehouse2014, 7780967, smith2020morphable, yang2020facescape, li2020learning, FLAME:SiggraphAsia2017, wang2022faceverse}. An alternative method is 3DMM built from a face image dataset~\cite{inproceedings, tran2018nonlinear}. The problem with the 3DMM-based approach is that it is difficult to represent the details of facial appearance using 3DMM. Other methods exist to represent facial shapes in more detail by adding geometric information~\cite{jackson2017large, guo2018cnnbased, 5432208, richardson2017learning, sela2017unrestricted, 8578975, tewari2018selfsupervised, 9321747}. However, it is still difficult to reconstruct a high-fidelity 3D face model using 3DMM fitting. For more detailed discussions on 3DMM, see~\cite{egger20203d}.

Appearance information is an essential factor in achieving high-fidelity 3D face reconstruction. Texture inference methods have been widely proposed for generating high-quality texture maps.
UV-GAN~\cite{deng2017uvgan} is a UV texture completion method. First, this method generates the face UV map texture by unwarping the input face image using the result of 3D face reconstruction. The generated face UV map has missing regions because of the occlusion. To complete these regions, an image-inpainting technique is applied. This UV map inpainting framework is widely extended ~\cite{Gecer_2021_CVPR, 9710579}.
Saito et al.~\cite{saito2017photorealistic} proposed a photorealistic face texture map generation method. They showed the possibility of a high-resolution face texture map by combining the low-frequency appearance information from 3DMM and the high-frequency appearance information from the input face image. Yamaguchi et al.~\cite{Yamaguchi:2018:ToG} employed image completion and image-to-image translation to generate diffuse, specular, and displacement texture maps. They also applied the super-resolution method~\cite{Ledig:2017:SRGAN} to obtain 2K-resolution texture maps. In the training process, they used various lighting conditions to render the synthetic face images to obtain the robustness of the lighting environment change. Using the generated texture maps, they can apply physically based rendering.
GANFIT~\cite{Gecer_2019_CVPR} uses the advantages of GAN and differentiable rendering to generate a face texture map from a latent vector and optimizes the entire network. They utilized losses based on face landmark detection and face recognition to maintain the identity and fit the face geometry. A similar approach was applied and improved by ~\cite{9156897}.
AvatarMe~\cite{Lattas_2020_CVPR} is an extension of GANFIT for generating high-fidelity texture maps. AvatarMe trains an image translation network to obtain textures for photorealistic rendering. AvatarMe can generate 4K-resolution diffuse albedo, diffuse normal, specular albedo, and specular normal texture maps. AvatarMe++~\cite{9606538} is an improvement of AvatarMe~\cite{Lattas_2020_CVPR}.
In contrast to the other methods, Bao et al.~\cite{Bao:2021:ToG} used an RGB-D selfie video input acquired by a consumer smartphone. They proposed a hybrid method of parametric fitting and CNN-based methods to estimate the reflectance. This method can produce albedo and normal texture maps.

All of the above appearance reconstruction methods directly use information from the input face image. Therefore, the texture generation process fails if the input images are taken under extreme conditions, such as lighting and expressions. To solve this problem, face normalization techniques~\cite{cole2017synthesizing, Nagano:2019:ToG, Luo_2021_CVPR} have gradually attracted attention, regarded as a preprocess of avatar creation.

\subsection{Face Normalization}
Many face normalization methods can be used for preprocessing to achieve robust and stable 3D face reconstruction.
Face-frontalization methods~\cite{huang2017face, towards-large-pose-face-frontalization-in-the-wild, yin2020dualattention} can correct face orientation to reduce the number of occluded areas. This process improves the accuracy of the 3D face reconstruction.
Lighting and shadow manipulation methods ~\cite{ 9010718, sfsnetSengupta18, 10.1145/3306346.3323008, zhang2020portrait, hou2021high, r2021photoapp, Pandey:2021, 10.1145/3414685.3417824, 10.1145/3386569.3392464} that adjust skin color can be further used to generate a stable albedo texture map.
Facial attribute editing methods~\cite{tewari2020stylerig, tewari2020pie, ghosh2020gif, Yang_2021_CVPR, lin2021anycost} are comprehensive tools that can change pose and lighting. In addition, these methods can restore the input facial expression to a neutral facial expression state.

Nagano et al.~\cite{Nagano:2019:ToG} proposed the first face normalization technique for generating an avatar.
Perspective distortion, lighting, head pose, and facial expressions in the input image were normalized through a step-by-step process. On the other hand, instead of using the input image directly, Luo et al.~\cite{Luo_2021_CVPR} employed StyleGAN2~\cite{karras2020analyzing} to refine the facial texture. Their method could generate normalized avatar face models under harsh lighting conditions.

Although various works can be used to handle lighting and expression for generating the 3D face model, to the best of our knowledge, no study has investigated the influence of facial makeup on high-fidelity 3D face reconstruction tasks. Recently, GAN-based face image generation can create photorealistic face images. In many cases, the output face image includes facial makeup, especially in female photos. Removing facial makeup is necessary to generate normalized avatar face models. Enlarging the scanned face dataset is not a realistic solution because there are numerous variations in makeup. 
Makeup transfer is a related research topic that can transfer makeup styles between facial images. 
BeautyGAN~\cite{Li:2018:MM} employed a histogram matching loss to maintain the color distribution between the face regions.
LADN~\cite{gu2019ladn} employed a local discriminator to enable a strong makeup style transfer.
PSGAN~\cite{jiang2019psgan} utilized an attention mechanism to accurately achieve makeup transfer between different head poses and facial expression images.
Nguyen et al.~\cite{m_Nguyen-etal-CVPR21} proposed a method to transfer makeup patterns precisely by unwarping the face image to a UV representation.
SCGAN~\cite{Deng_2021_CVPR} employed a style-based encoder to map the makeup style into a disentangled style code, which solves the large spatial misalignment problem. EleGANt~\cite{yang2022elegant} proposed a locally editable makeup transfer method that can achieve more flexible controls.

Unlike the above methods, our method does not require a reference image for makeup removal. In addition, for the application of high-fidelity texture inference, makeup-removed face images are preferably consistent with the high-resolution texture dataset that can achieve consistent normalized texture inference.
\section{Data Preparation}
We use two datasets. The makeup dataset was used to learn facial de-makeup. The high-resolution texture dataset was utilized to build the 3DMM and train the high-fidelity texture inference network.

\subsection{Makeup Dataset}
The current publicly available makeup dataset~\cite{Li:2018:MM, jiang2019psgan, gu2019ladn} is not ideal, because the dataset classified as non-makeup also contains many photos with light makeup. To efficiently remove the influence of makeup styles, we used the dataset created in the makeup transfer research LADN~\cite{gu2019ladn}, because they have strong contrasts. This dataset contains 334 faces without makeup and 355 faces with makeup. In addition, synthetic makeup images were generated by blending the makeup style and face images without makeup. For more details about the LADN dataset, please refer to \cite{gu2019ladn}.
Owing to the contribution of LADN, we can train our network via weakly supervised learning using before makeup and synthetic makeup image pairs.

\subsection{Scanned Face Dataset}
To prepare a high-resolution texture dataset, we captured 180 Japanese females with neutral facial expressions using the ESPER LightCage with 55 Sony $\alpha$7RIII cameras and polarizing lights.
A head mesh was reconstructed using the RealityCapture software.
Skin specular and diffuse components were separated by the cross-polarization technique using polarizing filters.
Normal maps were generated using a photometric stereo technique.
Finally, the head mesh and a set of diffuse, normal, roughness, and specular texture maps at 16K-resolution were obtained.
However, the raw scan data involved artifacts.
We asked the 3DCG artists to clean the scan data.
In addition, the reconstructed head mesh was registered with the same topology during the clean-up process.

\subsection{3DMM Construction}
We used the scanned head meshes and diffuse texture maps to create a linear PCA-based 3DMM~\cite{10.1145/311535.311556}. To improve the representation of the expression, we manually made 236 blendshapes from the mean head.
The constructed 3DMM contained 65,143 vertices and 130,000 faces. 
The shape $S$ and appearance $A$ of the 3DMM can be controlled by changing the parameters of identity $\alpha$, expression $\beta$, and appearance $\delta$.
\begin{equation}
\begin{aligned}
&S = \bar{S} + B_{id}\alpha + B_{exp}\beta\\
&A = \bar{A} + B_{a}\delta
\end{aligned}
\end{equation}
where $\bar{S}$ and $\bar{A}$ are the mean shape and appearance, respectively.
$B_{id}$, $B_{exp}$, and $B_{a}$ are the identity basis, expression basis, and
appearance basis vectors, respectively.
In our method, we employ ~\cite{deng2020accurate} as the backbone of our neural network to be optimized and regress the 3DMM coefficients $C(\alpha, \beta, \delta, R, t)$.
The coefficient vector $C\in\mathbb{R}^{298}$ was composed of the parameters of shape identity $\alpha\in\mathbb{R}^{120}$, expression $\beta\in\mathbb{R}^{120}$, appearance $\delta\in\mathbb{R}^{52}$, rotation $R\in\mathbb{R}^{3}$, and translation $t\in\mathbb{R}^{3}$. In addition, spherical harmonics (SH) lighting is parameterized $\gamma\in\mathbb{R}^{27}$ in the case of lighting conditions.

We use two 3D face reconstruction networks $F$ and $E$. We employ ResNet50~\cite{7780459} as the backbone of network $F$ trained with our 3DMM following \cite{deng2020accurate} to estimate $\textit{C}$ and SH lighting. The network $E$ is ResNet18~\cite{7780459} architecture to estimate $\hat{\textit{C}}$.
\section{BareSkinNet}

Given a makeup face image, the de-makeup and de-lighting network (BareSkinNet) generates a bare skin image with a 3D facial geometry and coarse 3D textures.

The training process of BareSkinNet is shown in Fig. \ref{fig:01_method}. The framework comprises two parts. 1) De-makeup and de-lighting network $G$ for removing makeup and lighting influences from the input image to generate a bare skin image (Sec. \ref{sec:demake_delight}). 2) 3D face reconstruction network $E$ and $F$ for estimating the 3DMM coefficients and SH lighting (Sec. \ref{sec:3dmmfitting}). The coefficients and rendered results are used to improve the capability of $G$.
The loss function for BareSkinNet is represented as follows:
\begin{equation}
L_{BSN} = L_{DD} + L_{FR}
\end{equation}
$L_{DD}$ is a loss function for the de-makeup and de-lighting network, and $L_{FR}$ is a loss function for the 3D face reconstruction process.
The details of each loss are described in the following sections.

\subsection{De-makeup and De-lighting network}
\label{sec:demake_delight}

\indent We employ a U-Net ~\cite{ronneberger2015unet} architecture with skip connections for de-makeup and de-lighting network $G$ to maintain the structure of the input face image. Using this architecture, we can remove makeup and lighting influences from the input face image while maintaining the identity information.

As shown in Fig. \ref{fig:01_method}, we used the pairs of before and after makeup images from the LADN makeup dataset~\cite{gu2019ladn}. We remove makeup by minimizing the distance between the bare skin image and the before makeup face image. The loss function for the de-makeup and de-lighting is defined as follows: 

\begin{equation}
L_{DD} = w_1L_{photo} + w_2L_{GAN} + w_3L_{LPIPS}
\end{equation}

$L_{photo}$ is the L1 pixel loss, $L_{GAN}$ is the adversarial loss calculated by PatchGAN~\cite{isola2017image} for realistic results, and $L_{LPIPS}$ is the perceptual image patch similarity (LPIPS) metric loss~\cite{zhang2018perceptual} preserved meaningful facial features. $w_1$, $w_2$, and $w_3$ are the weights for balancing each term. These losses are calculated from the bare skin image and the corresponding reference image.

The makeup loss proposed in BeautyGAN~\cite{Li:2018:MM} has been widely used in makeup transfer tasks. In contrast to the makeup transfer task, we do not need to ensure consistency between the color distributions of different makeup styles. In addition, we remove the lighting influence in our framework by minimizing loss with 3D face reconstruction. For these reasons, we do not employ makeup loss. Note that the purpose of this network is to remove the makeup and lighting influences from the input image. However, the face image before the makeup used as the ground truth already contains lighting information. Therefore, it is difficult to remove the lighting influence using this network completely. We verified this claim in an ablation study (Sec. \ref{sec:ablation_study}). To remove the influence of lighting, we incorporate 3D face reconstruction networks. In the next section, we discuss using the 3D face reconstruction process to achieve de-lighting.
    
\subsection{De-lighting via 3D face reconstruction}
\label{sec:3dmmfitting}

\indent To remove the lighting influence, we use 3D face reconstruction networks $E$ and $F$ to estimate the 3DMM coefficients and SH lighting from the bare skin image and reference image.

Since the LADN dataset~\cite{gu2019ladn} only has a small number of subjects, the pre-trained network $F$ will be fixed as a teacher role to help the network $E$ learn the 3D face reconstruction process. 
We expect that jointly learning the 3D face reconstruction process will improve the capability of network $G$ to remove makeup and lighting influences.
In the training stage, we jointly learn networks $G$ and $E$ with fixed $F$. We employ differentiable rendering to render the 3DMM so that BareSkinNet can be trained in an end-to-end fashion. Therefore, the networks $G$ and $E$ can be optimized simultaneously.
In the inference stage, we use the network $G$  to obtain the bare skin image, and then use $F$ to obtain 3DMM coefficients. Because $F$ is trained from a large dataset, which can perform a more accurate 3D reconstruction.

For the differentiable renderer $DR$, we use Nvdiffrast~\cite{Laine2020diffrast} to calculate the reconstruction loss and optimize the parameters of the network. 
$\gamma$ is estimated from the pre-makeup reference image by $F$. The consistency loss between $E$ and $F$ results implicitly uses estimated SH lighting.
After estimating 3DMM coefficients $\hat{\textit{C}}$ from $E$ and SH lighting from $F$, we render the 3DMM with and without lighting. Note that \textit{C} is not used for 3DMM rendering. The rendered 3DMM image without lighting should be close to the bare skin image, and the rendered 3DMM image with lighting should be close to the reference image. 
By combining with the de-makeup and de-lighting network $G$, BareSkinNet can correctly remove the makeup and lighting influences.
The following loss function is used for the 3D face reconstruction process to optimize network $E$.

\begin{equation}
\begin{split}
L_{FR} &= w_{coeff}L_{coeff} + w_{land}L_{land} \\
& + w_{diff}L_{diff} + w_{light}L_{light} + w_{reg}L_{reg}
\end{split}
\end{equation}

$L_{coeff}$ is the loss between 3DMM coefficients $\hat{\textit{C}}$ and $\textit{C}$ estimated from networks $E$ and $F$.
$L_{land}$ is the reprojection error of facial landmarks between the detected 2D landmarks from the reference image and the projected landmarks from the 3DMM. $L_{diff}$ and $L_{light}$ are the pixel-wise L1 distances between the images and the rendered 3DMM image with and without lighting, respectively. $L_{reg}$ is the regularization term for the coefficients of the 3DMM. This regularization term is commonly used in the 3DMM-based face reconstruction process to avoid an unnatural face output. Concretely, $L_{reg}$ is defined as the distance between the estimated and mean face coefficients. $w_{coeff}$, $w_{land}$, $w_{diff}$, $w_{light}$ and $w_{reg}$ are weights for balancing each term.

By combining the process of 3D face reconstruction, the lighting influence can be removed in the bare skin image, and the overall skin tone of the face region is matched with the 3DMM diffuse. The 3DMM diffuse retains global appearance information at a low frequency in the high-resolution texture dataset.

\section{High-fidelity texture inference}
This process is considered to be an application. We can apply high-fidelity texture inference methods on top of BareSkinNet. Similar to that reported in the literature~\cite{Yamaguchi:2018:ToG, li2020learning, Lattas_2020_CVPR, 9606538}.

The high-fidelity texture inference network takes the bare skin image and the 3D face reconstruction result acquired by BareSkinNet as shown in Fig. \ref{fig:01_method}. The bare skin image, 3DMM diffuse, 3DMM normal, and 3DMM geometry are unwarped to the same UV map. Then, the image-translation framework~\cite{wang2018pix2pixHD} can infer diffuse, normal, roughness, and specular texture maps. The output of the image-translation framework is 1K-resolution. Finally, we use SRGAN~\cite{Ledig:2017:SRGAN} to upscale the 1K-resolution texture maps to 4K-resolution texture maps.

\begin{figure}[t]
    \centering
    \includegraphics[width=\linewidth]{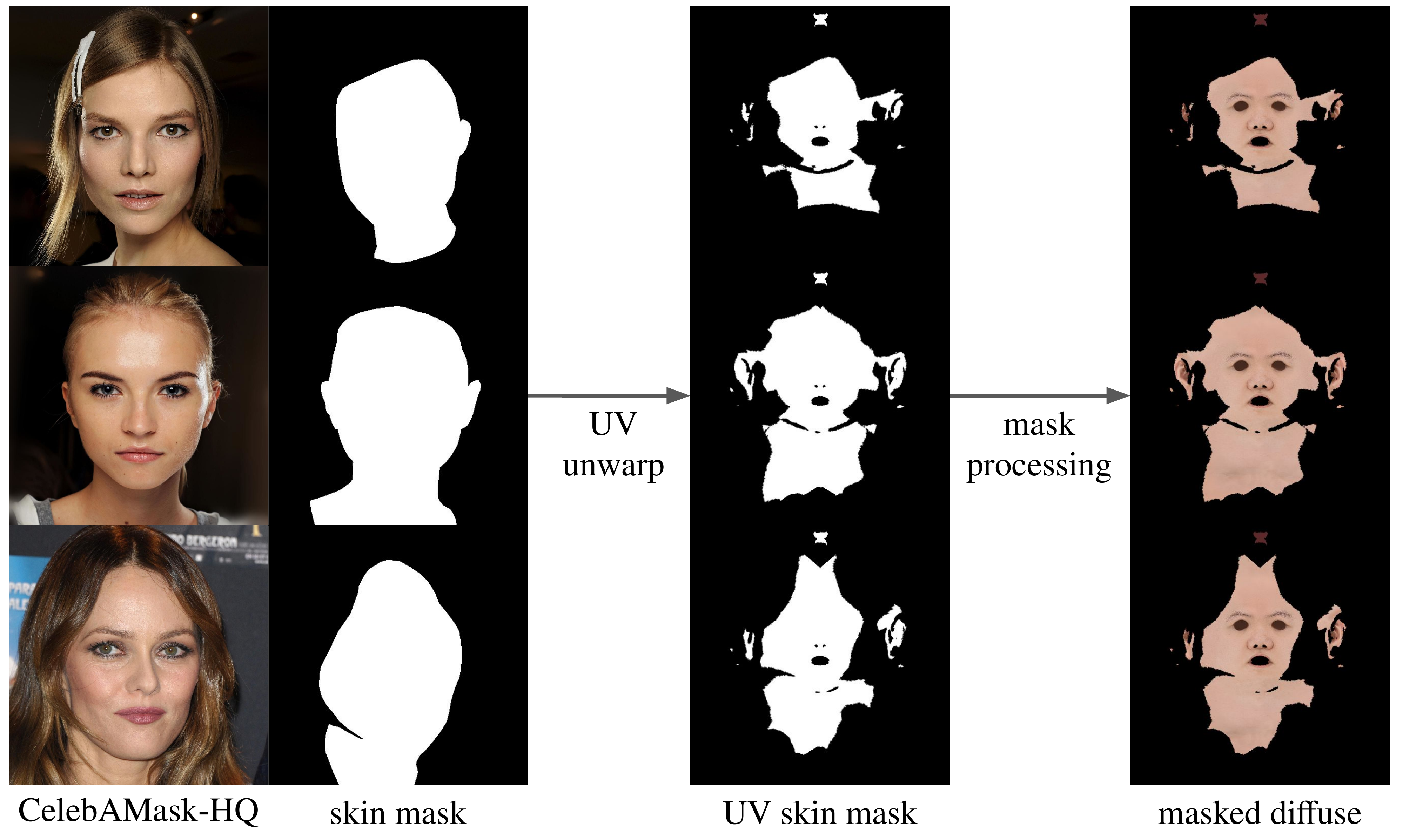}
    \caption{Data augmentation using 2D skin masks. Skin masks are unwarped using the result of 3D face reconstruction. These unwarped skin masks are used to synthesize the occlusion effect.} 
\label{fig:04_method}
\end{figure}

In the training process, to synthesize occlusions depending on the viewing angles, we rendered a scanned face model with diffuse from three different viewpoints: front, left, and right. The 3DMM is then fitted to the rendered image. Following the 3D face reconstruction result, the rendered image is unwarped to the UV map, which is the same as the inference process. As a result of this process, we can obtain the pairs of the occluded high-frequency appearance information of face texture, the low-frequency appearance information of 3DMM, geometry, and normal, which correspond to high-resolution diffuse, normal, specular, and roughness texture maps.
In addition, to improve the robustness against occlusion due to viewpoint change, we synthesize the occluding effect using 2D face masks as shown in Fig. \ref{fig:04_method}. First, we carefully selected face images and corresponding skin mask images for 500 neutral faces from the celebAMask-HQ dataset~\cite{CelebAMask-HQ}, and the 3DMM was fitted to face images. Using the result of 3D face reconstruction, skin mask images were unwarped to the UV map. In each training process, unwarped visible masks were used to enhance the capacity to handle occlusion by multiplying it with the face texture.  

\section{Implementation detail}
We implemented our pipeline with PyTorch. For the differentiable renderer, we use Nvdiffrast~\cite{Laine2020diffrast}. All training processes were performed on an NVIDIA RTX 2080Ti graphics card.
The 3DMM was created using the scanned face dataset. In the de-makeup and de-lighting stage, the face images from the LADN dataset were resized to $256 \times 256$. 3D face reconstruction was performed under the same resolution and finally rendered to a $1024 \times 1024$ UV map representation. 
In the high-fidelity texture inference stage, the high-resolution texture contained in the scanned face dataset is resized to 4K-resolution. SRGAN is also trained under $4\times$ super-resolution conditions to upscale the image from 1K-resolution to obtain the final 4K-resolution result.

We set our balancing factors as the following: $w_1 = 100$, $w_2 = 1$,  $w_3 = 40$, $w_{coeff} = 1\mathrm{e}{-1}$, $w_{land} = 8\mathrm{e}{-2}$, $w_{diff} = 100$, $w_{light} = 100$,$w_{reg} = 1\mathrm{e}{-3}$. First, The BareSkinNet was trained for 10000 iterations only using $L_{DD}$. This can warm up the BareSkinNet and generate a stable bare skin image. At this stage, the BareSkinNet is split and only $G$ is used. Then the BareSkinNet was trained in 30000 iterations with the full model. We set a batch size of 4 using the Adam optimizer to train our BareSkineNet. 
The learning rate of the de-makeup and de-lighting network $G$ was set to $2\mathrm{e}{-5}$ and exponential decay rates ($\beta1$, $\beta2$) = (0.5, 0.999). 
The learning rate of the 3D face reconstruction network $E$ was set to $1\mathrm{e}{-4}$.
\section{Results and Evaluations}

To demonstrate the effectiveness of our method, we conducted qualitative and quantitative evaluations. In the qualitative evaluation, we present the results of the bare skin images. In addition, we show the rendering results using the generated high-fidelity texture maps. By comparing it with the existing high-fidelity texture inference method~\cite{Yamaguchi:2018:ToG}\footnote{We used the original implementation and pre-trained model provided by the authors.} and the normalized avatar synthesis method~\cite{Luo_2021_CVPR}\footnote{Results were provided by the authors.}, we demonstrate the usefulness of our method against makeup. In the quantitative evaluation, we evaluated the stability of the generated texture maps.

\subsection{Qualitative Evaluation}

\
\newline
\noindent\textbf{De-makeup and de-lighting}

\begin{figure*}[t]
    \centering
    \includegraphics[width=\linewidth]{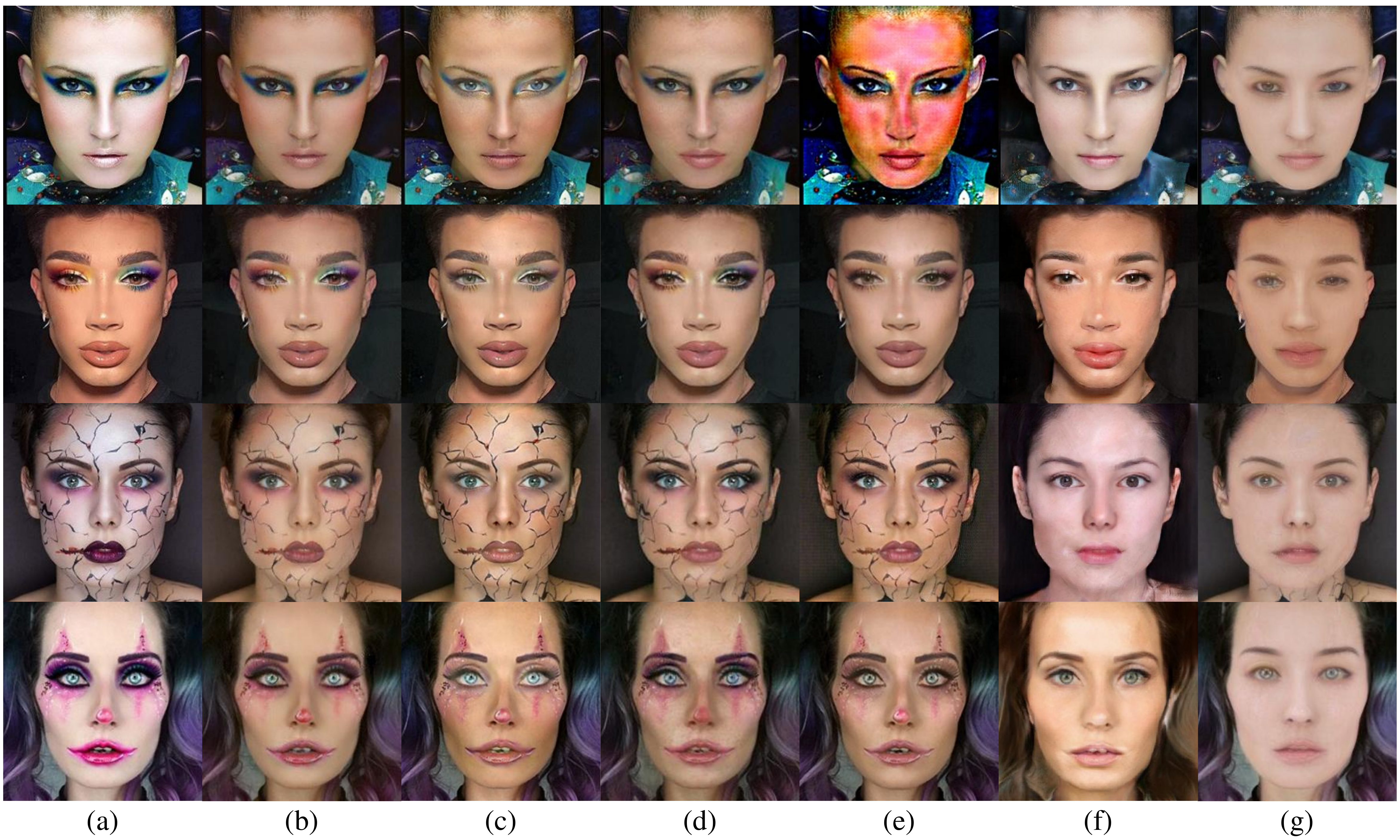}
    \caption{Comparison with state-of-the-art makeup transfer methods for makeup removal. From left to right, we show (a) input face images; (b) BeautyGAN~\cite{Li:2018:MM};  (c) PSGAN~\cite{jiang2019psgan}; (d) SCGAN~\cite{Deng_2021_CVPR}; (e) EleGANt~\cite{yang2022elegant}; (f) results from the paper of LADN~\cite{gu2019ladn}; (g) our results of bare skin images.}
\label{fig:demake_result}
\end{figure*}

First, we compared the makeup removal effects with state-of-the-art methods\cite{Li:2018:MM, jiang2019psgan, Deng_2021_CVPR, yang2022elegant, gu2019ladn}. As shown in Fig. \ref{fig:demake_result}, our BareSkinNet can remove the makeup and lighting influences without specifying a reference image or a known lighting condition. The state-of-the-art makeup transfer methods could not achieve makeup removal successfully. LADN~\cite{gu2019ladn} could remove makeup to some extent by carefully selecting a non-makeup reference image. However, the results were affected by the reference image and introduced new lighting. We dig deeper into the reasons for BareSkinNet effectiveness. We think the result of 3D face reconstruction can be regarded as a reference image. The diffuse of 3DMM contains ideal conditions without makeup and lighting. 
Additionally, the 3D face reconstruction process has consistency with the original image. Therefore, in contrast to using another person's reference image, our results preserve the subject's identity.

\begin{figure}[t]
    \centering
    \includegraphics[width=\linewidth]{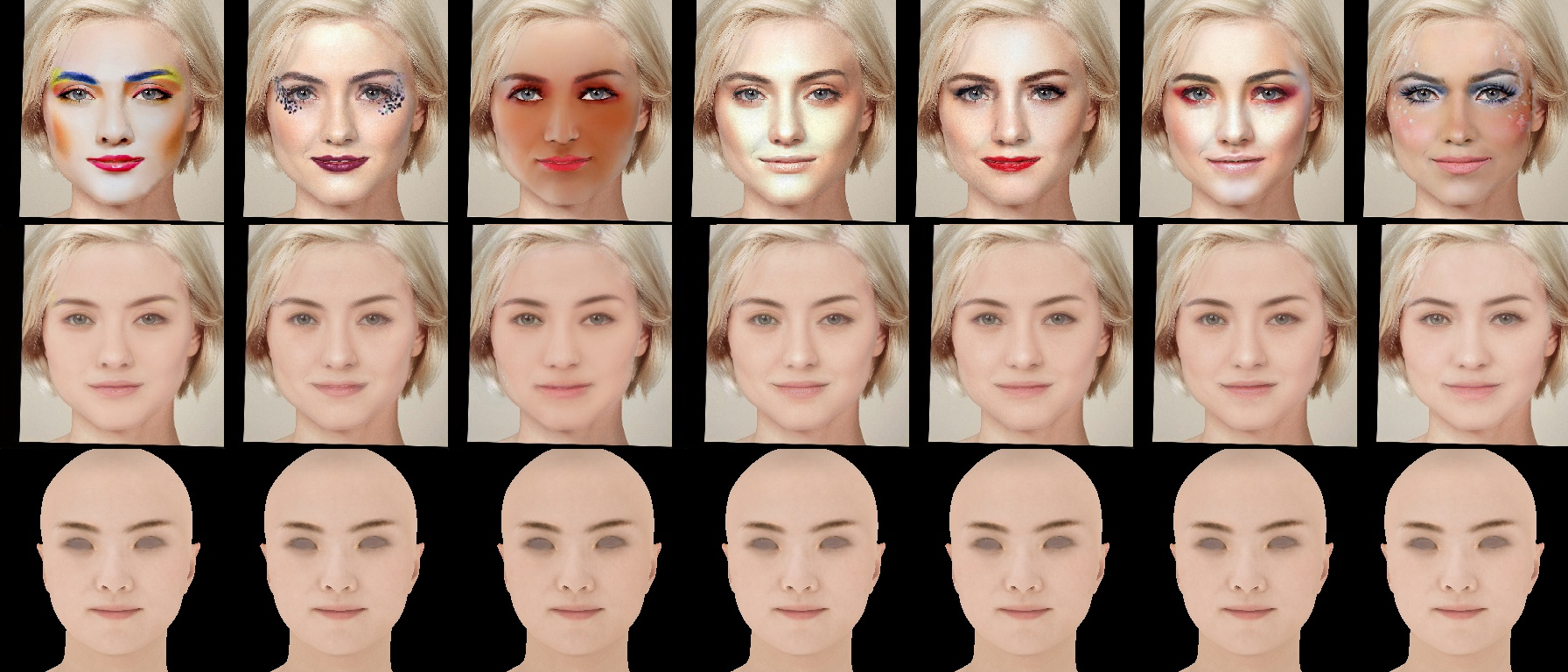}
    \caption{BareSkinNet for the same subject. The first row is the input images. The second row is the bare skin images. The last row is the reconstructed 3DMM from the bare skin images.} 
\label{fig:02_result}
\end{figure}

\begin{figure}[t]
    \centering
    \includegraphics[width=\linewidth]{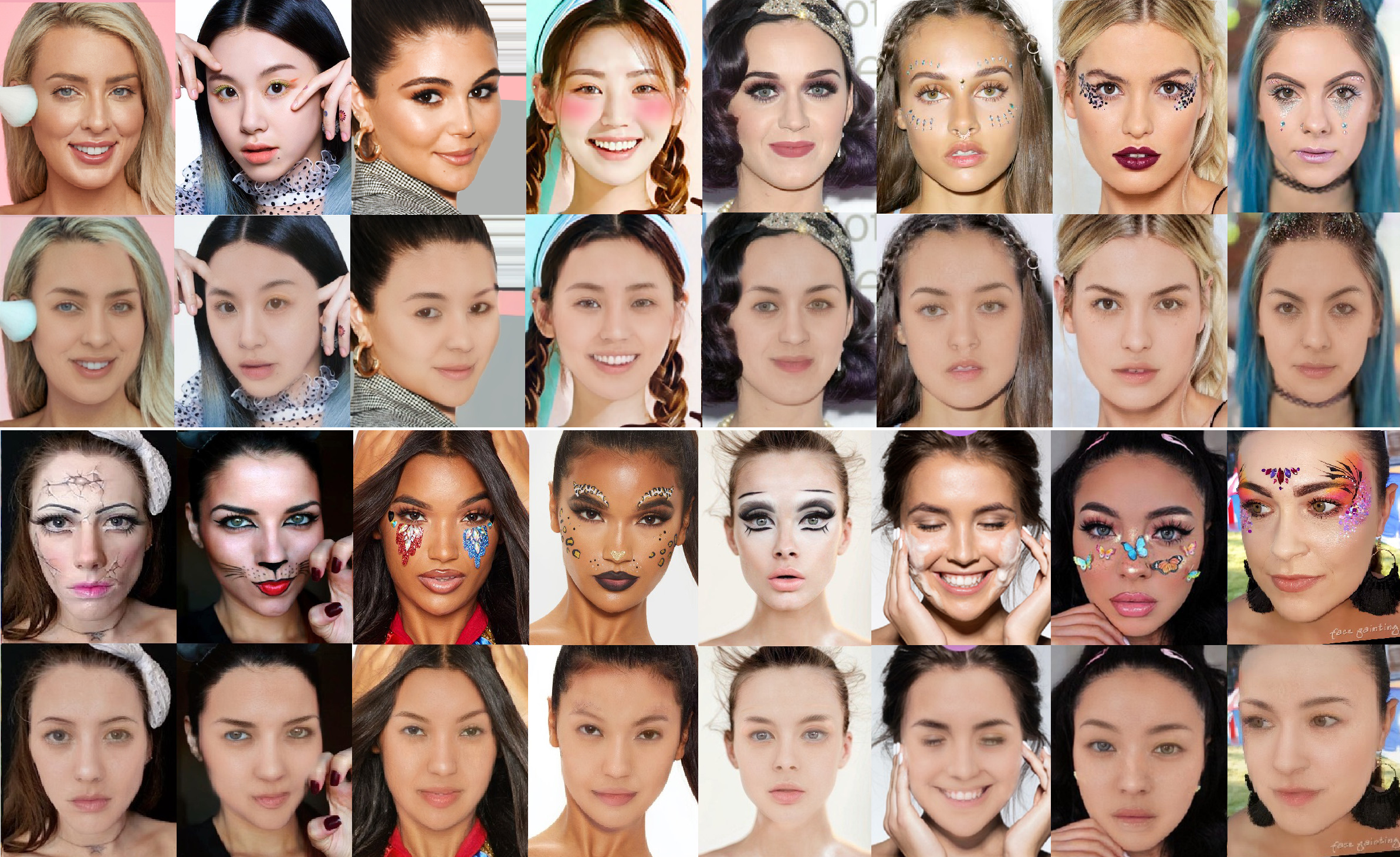}
    \caption{Results for real-world makeup image from CPM-Real dataset~\cite{m_Nguyen-etal-CVPR21}. The first and third rows are the input face images (light makeup and strong makeup). The second and the fourth rows are the bare skin images using BareSkinNet.} 
\label{fig:03_result}
\end{figure}
\begin{figure}[t]
    \centering
    \begin{subfigure}[htb]{0.12\linewidth}
        \centering
        \includegraphics[width=\linewidth]{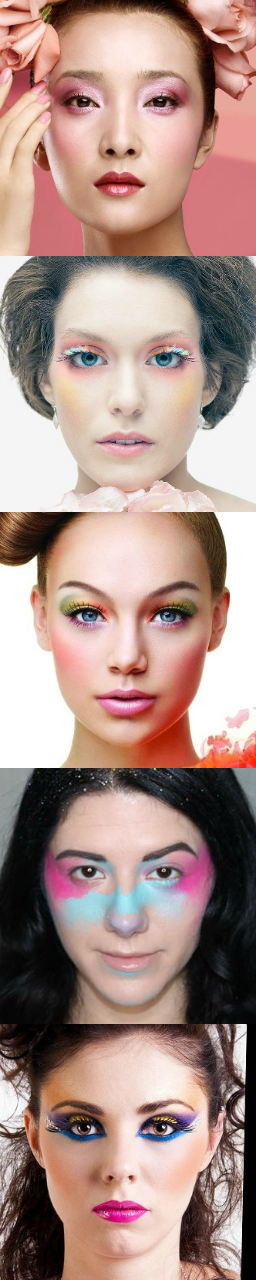}
        \caption{}
        \label{fig:01_a_result}
    \end{subfigure}
    \hspace{-0.2cm}
    \begin{subfigure}[htb]{0.12\linewidth}
        \centering
        \includegraphics[width=\linewidth]{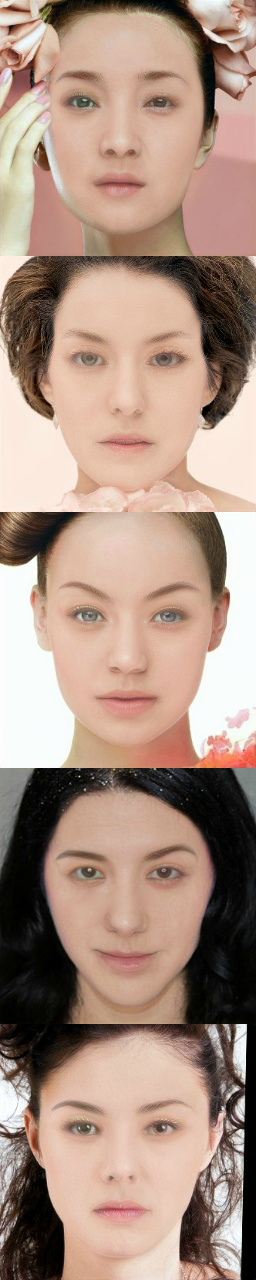}
        \caption{}
        \label{fig:01_b_result}
    \end{subfigure}
    \hspace{-0.2cm}    
    \begin{subfigure}[htb]{0.12\linewidth}
        \centering
        \includegraphics[width=\linewidth]{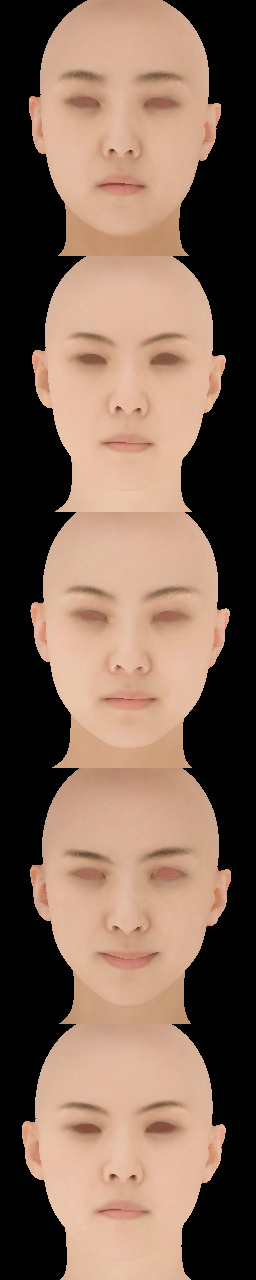}
        \caption{}
        \label{fig:01_c_result}
    \end{subfigure}
    \hspace{-0.2cm} 
    \begin{subfigure}[htb]{0.48\linewidth}
        \centering
        \includegraphics[width=\linewidth]{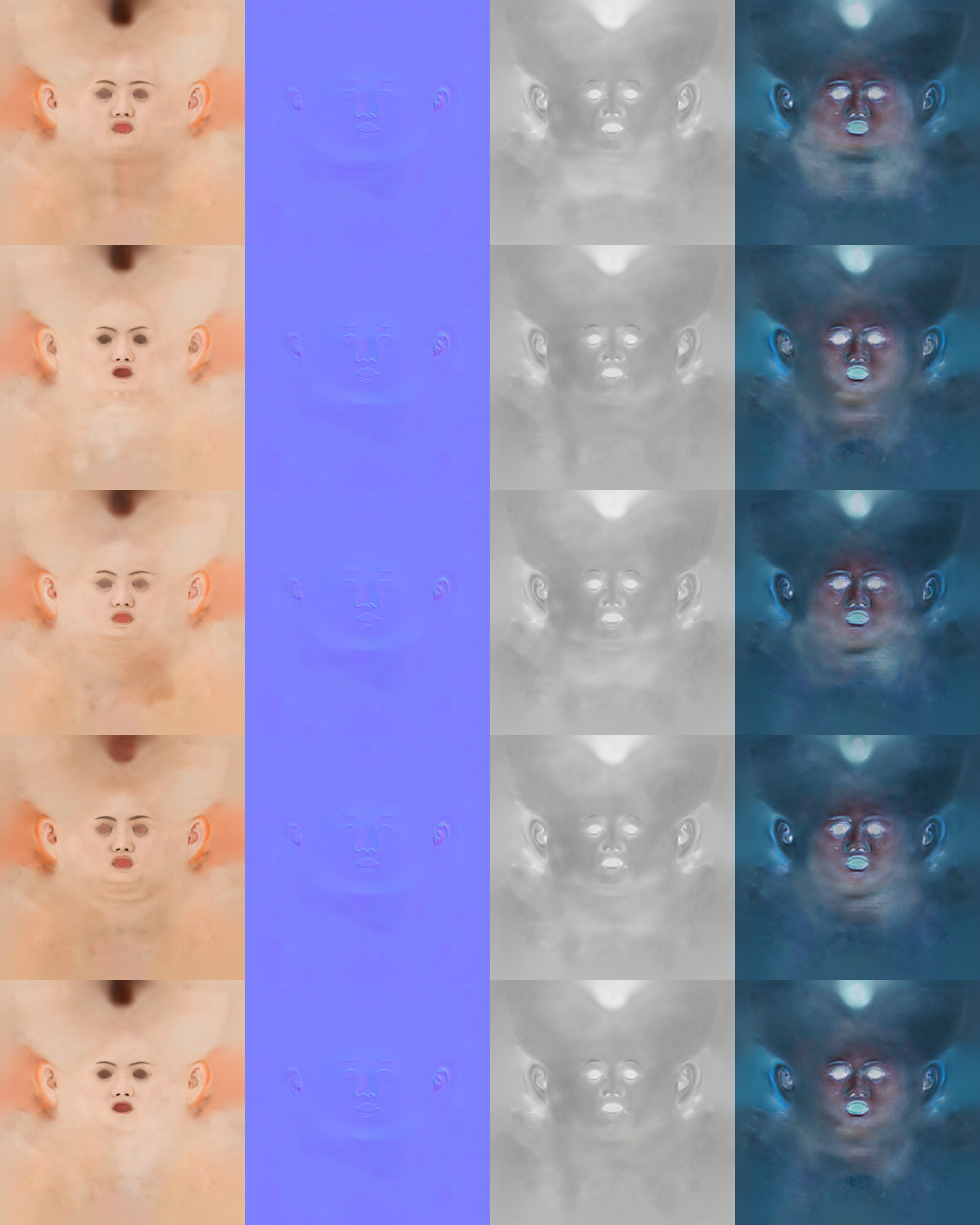}
        \caption{}
        \label{fig:01_d_result}
    \end{subfigure}
    \hspace{-0.2cm} 
    \begin{subfigure}[htb]{0.12\linewidth}
        \centering
        \includegraphics[width=\linewidth]{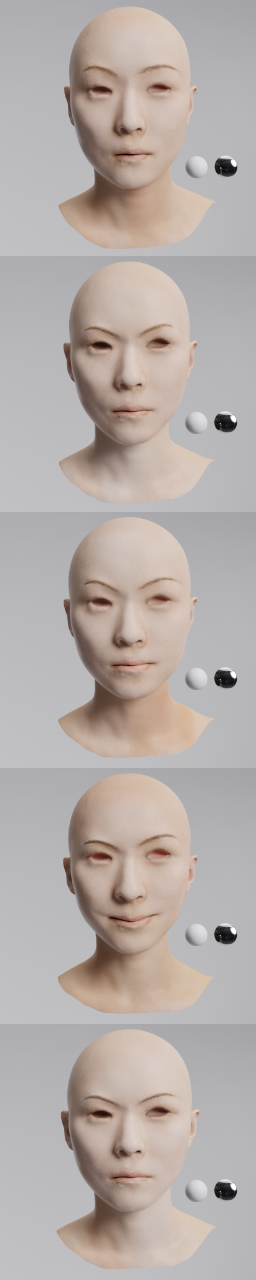}
        \caption{}
        \label{fig:01_e_result}
    \end{subfigure}
    \hspace{-0.2cm} 
    \caption{Example results of normalized texture inference. From left to right, we show (a) makeup images; (b) recovered bare skin images; (c) 3D face reconstruction results; (d) inferred diffuse, normal, roughness and specular texture maps; (e) rendering results using inferred texture maps.}
    \label{fig:01_result}
\end{figure}
\begin{figure}[t]
    \centering
    \includegraphics[width=\linewidth]{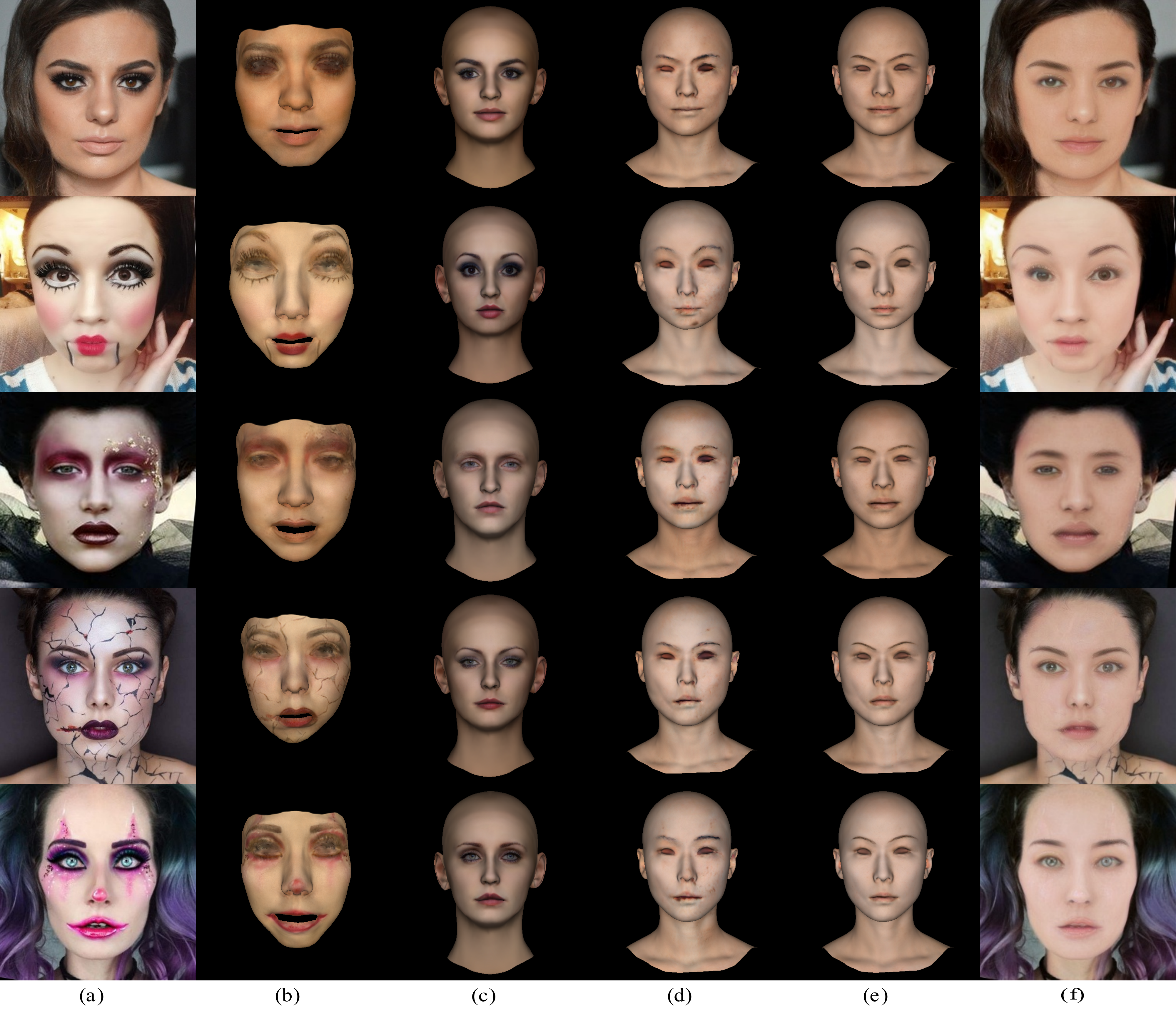}
    \caption{Qualitative comparison with state-of-the-art methods. From left to right, we present (a) input images; (b) results of Yamaguchi et al. \cite{Yamaguchi:2018:ToG}; (c) Luo et al. \cite{Luo_2021_CVPR}; (d) ours without BareSkinNet; (e) ours; (f) output of BareSkinNet. Result images are rendered using inferred diffuse texture.} 
\label{fig:04_result}
\end{figure}

\begin{figure*}[t]
    \centering
    \includegraphics[width=\linewidth]{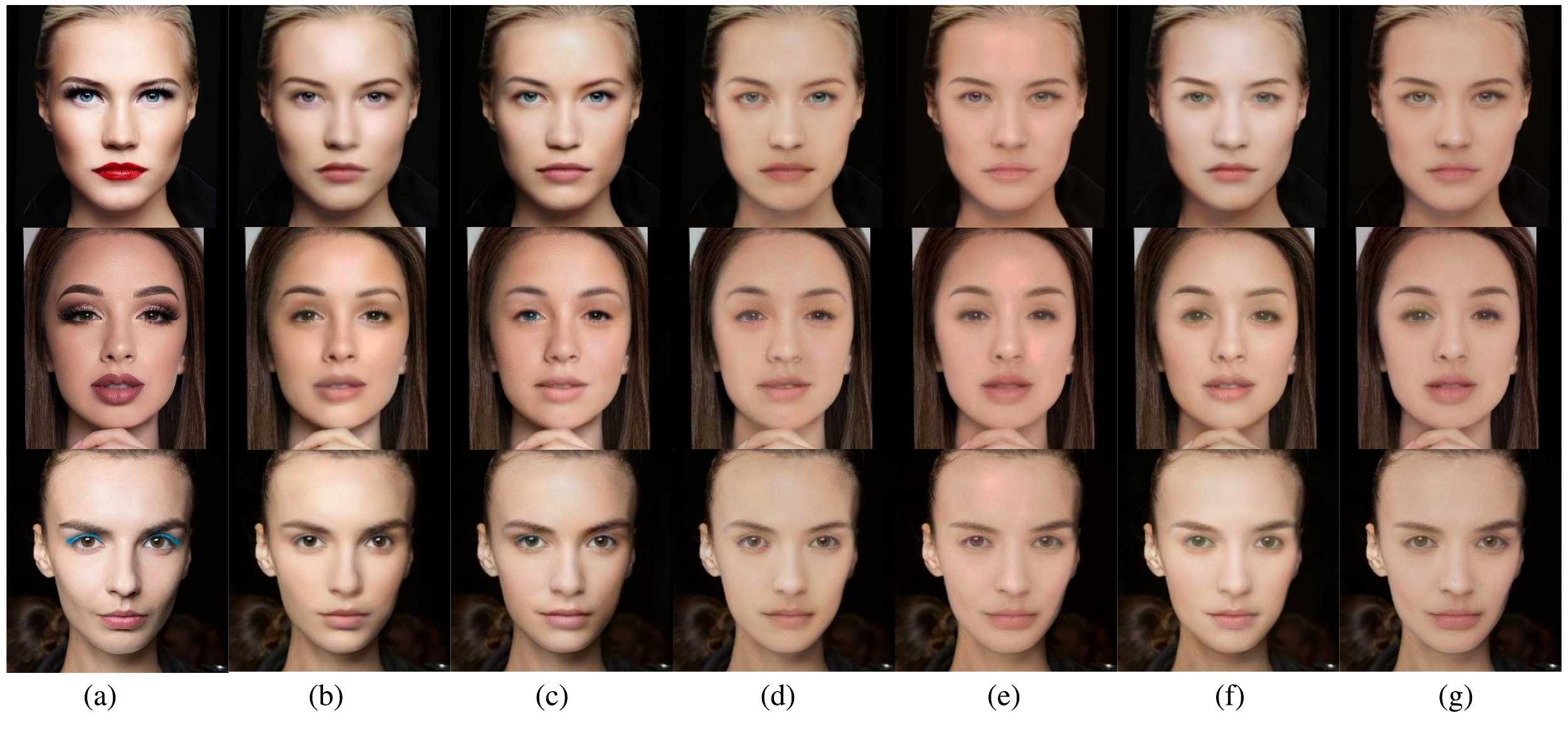}
    \vspace{-7mm}
    \caption{Results of ablation study. (a) input images; (b) $L_{photo}$ and $L_{GAN}$; (c) $L_{photo}$, $L_{GAN}$ and $L_{LPIPS}$; (d) use $F$ instead of $E$ without fine-tuning; (e) use $F$ instead of $E$ with fine-tuning; (f) our proposed teacher-student framework without $L_{light}$; (g) our full model.} 
\label{fig:05_result}
\end{figure*}

BareSkinNet can remove makeup and lighting influences correctly. Fig. \ref{fig:02_result} shows the results of BareSkinNet for the same subject. These results confirm that our method can produce a consistently clean face under different makeup and lighting conditions.

Fig. \ref{fig:03_result} shows the other results of the de-makeup and de-lighting. 
The makeup face images were obtained from the CPM-Real dataset~\cite{m_Nguyen-etal-CVPR21}. The CPM-Real dataset contains real-world makeup photos. From these results, our method can successfully remove the makeup and lighting influences from the light makeup images. In addition, even for the strong makeup images, the results of de-makeup and de-lighting are of reasonable quality.

\
\newline
\noindent\textbf{Texture inference}

Fig. \ref{fig:01_result} shows the results of the normalized texture inference. We selected five samples of face images from the LADN dataset. The makeup face image was used as an input to BareSkinNet. BareSkinNet outputs a bare skin image and 3D face reconstruction result. The results confirm that BareSkinNet can successfully remove the makeup and lighting influence from the input image. The high-fidelity texture inference network was then executed using the outputs of BareSkinNet. We can confirm that the generated clean texture maps can be used for realistic face rendering.

To demonstrate the effectiveness of BareSkinNet for high-fidelity texture inference, we compared the output texture maps with and without BareSkinNet preprocessing. As input to the texture inference process, with and without of BareSkinNet samples colors from the makeup image and the bare skin image, respectively. The entire 3D reconstruction process is the same except for the color sampling.

Figs. \ref{fig:04_result} (d) and (e) show the rendered results using the inferred diffuse texture maps with inputs (a) and (f), respectively. We can confirm artifacts in the generated diffuse texture maps in the case without BareSkinNet. On the other hand, using the BareSkinNet output, the generated diffuse texture maps do not include any artifacts.

We also compared state-of-the-art texture inference methods~\cite{Yamaguchi:2018:ToG, Luo_2021_CVPR}, as shown in Fig. \ref{fig:04_result} (b) and (c). 
For comparison, we only used the diffuse texture map for rendering. In the results of Yamaguchi et al.~\cite{Yamaguchi:2018:ToG}, the entire makeup texture patterns remained in the inferred diffuse texture maps. Although Luo et al.~\cite{Luo_2021_CVPR} could remove some makeup texture patterns, makeup effects remain around the eyebrow, eyes, and mouth. In contrast to these methods, our method successfully removed the makeup patterns for the entire face texture.

\subsection{Quantitative Evaluation}
Since it is hard to acquire ground truth data of in-the-wild de-makeup and de-lighting images, it is challenging to evaluate BareSkinNet directly. 
We evaluated the effectiveness of BareSkinNet by comparing the final outputs of the texture maps from the texture inference network. We randomly selected 70 subject faces with various synthetic 2100 makeup images that were not included in the training set. For comparison, we obtained the final texture maps with and without BareSkinNet. We used the output texture map of the before-makeup image inputs as a reference for computing the error metrics. We then computed the root mean square error (RMSE), peak signal-to-noise ratio (PSNR), and structural similarity index measure (SSIM) as metrics between the generated UV maps. Table~\ref{tab:table_1} lists the results for each score for each texture map. Compared with the texture maps without BareSkinNet, the errors of RMSE were reduced after applying BareSkinNet. In addition, the results confirm that the PSNR and SSIM metrics are improved. From these results, we believe that our BareSkinNet can improve the stability of 3D face reconstruction under various makeup and lighting conditions.

\begin{table}[t]
\caption{Quantitative evaluation of the output texture maps from the facial details inference network. Metrics are root mean square error (RMSE), peak signal-to-ratio (PSNR), and structural similarity index measure (SSIM).}
\label{tab:table_1}
\begin{center}
\begin{tabular}{l|cc}
\toprule
& w/o BareSkinNet & w/ BareSkinNet\\
\hline
\midrule
RMSE (Diffuse)& 3.910& \textbf{3.358}$\downarrow$\\
RMSE (Normal)& 1.365& \textbf{1.244}$\downarrow$\\
RMSE (Specular)& 3.997& \textbf{3.146}$\downarrow$\\
RMSE (Roughness)& 1.782& \textbf{1.242}$\downarrow$\\
\hline
PSNR (Diffuse)& 33.332& \textbf{36.167}$\uparrow$\\
PSNR (Normal)& 45.541& \textbf{46.252}$\uparrow$\\
PSNR (Specular)& 35.188& \textbf{37.843}$\uparrow$\\
PSNR (Roughness)& 43.850& \textbf{46.296}$\uparrow$\\
\hline
SSIM (Diffuse)& 0.963& \textbf{0.969}$\uparrow$\\
SSIM (Normal)& 0.976& \textbf{0.980}$\uparrow$\\
SSIM (Specular)& 0.943& \textbf{0.961}$\uparrow$\\
SSIM (Roughness)& 0.981& \textbf{0.985}$\uparrow$\\
\bottomrule
\end{tabular}
\setlength{\belowcaptionskip}{10pt}
\end{center}
\end{table}


\subsection{Ablation Study}
\label{sec:ablation_study}

To validate the effectiveness of each component in BareSkinNet, we conducted experiments with different submodules and losses. Example results are  presented in Fig. \ref{fig:05_result}. 

First, we only use de-makeup and de-lighting network $G$ results in supervised learning. As shown in (b) and (c), makeup effects were removed in most parts. However, some effects remain around the eyes. Utilizing $L_{LPIPS}$ can add some details and make the face clearer. 

Next, we added a pre-trained 3D face reconstruction network $F$. We verified the effectiveness of removing lighting and makeup with and without fine-tuning. As shown in (d) and (e), makeup around the eyes is better removed. Different from (b) and (c), we can confirm that the lighting effect is removed. However, some artifacts appeared in (e) around the illuminated area.

Finally, we went a step further to improve the capability of removing makeup and lighting influences by adding a network $E$ that can be trained. We found that the best performance can be achieved by letting network $E$ learn the diffuse part while the pre-trained network $F$ provides accurate SH lighting estimates. In addition, network $E$ is difficult to train without the support of a teacher-student strategy due to the limitation of the diversity of the makeup dataset. Comparing (f) and (g), by adding $L_{light}$, the lighting effect was removed significantly.
\section{Limitations}
\begin{figure}[t]
    \centering
    \includegraphics[width=\linewidth]{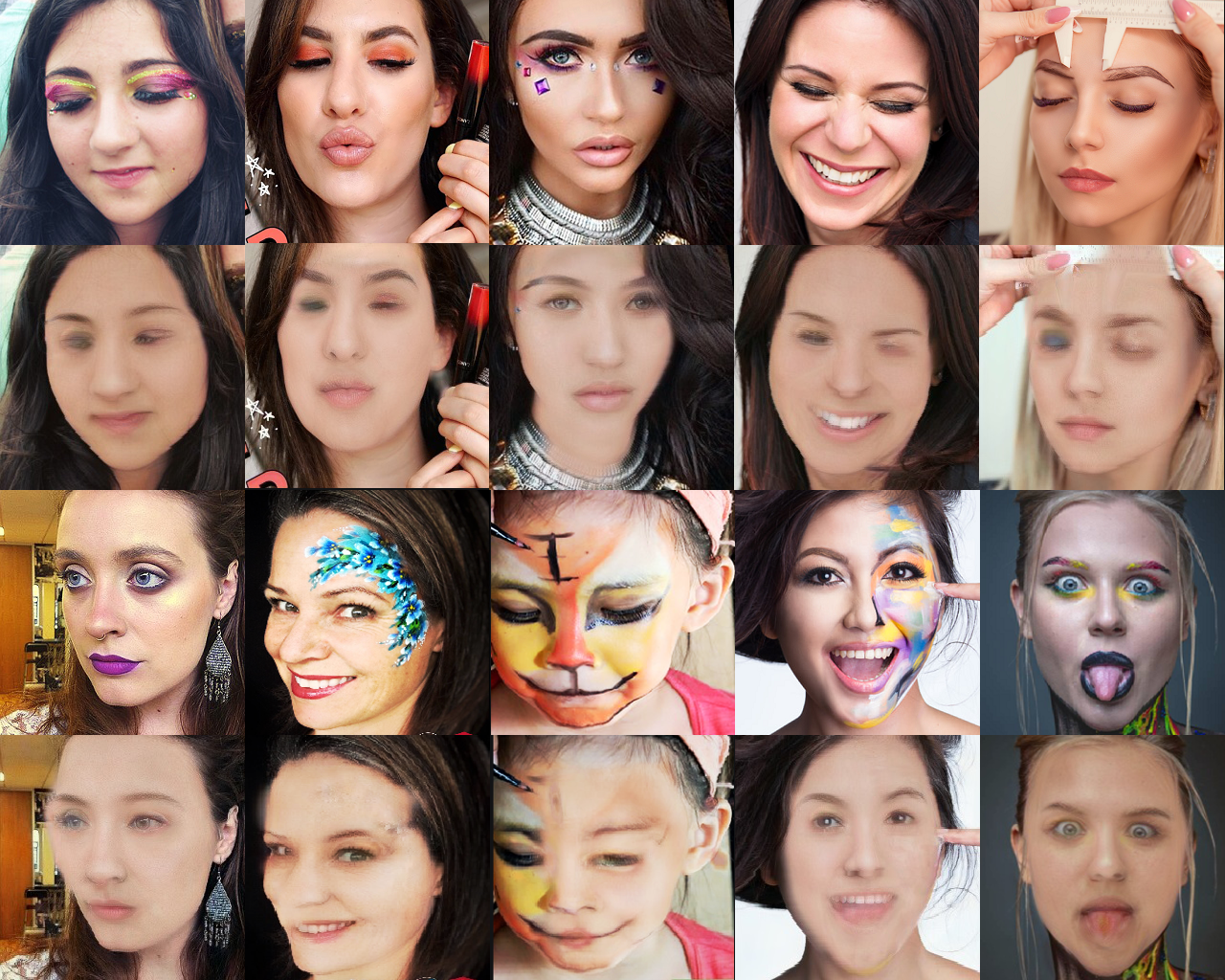}
    \caption{Examples of the failure cases of de-makeup}
    \vspace{-3mm}
\label{fig:07_limitation}
\end{figure}

Although the results of qualitative and quantitative experiments confirm that our method performed excellently under makeup and lighting conditions, it is challenging to handle largely inclined face poses and extreme facial expressions, especially a face with closed eyes or opened mouth.
Fig. \ref{fig:07_limitation} shows examples of the failure cases. BareSkinNet cannot produce desirable makeup removal results because BareSkinNet was trained using a front neutral expression face collection dataset. 

Also, In the makeup removal process, eyes and hair are affected by the color of 3DMM, which is a limitation. Although hair color and eye color changed, the result of the application in this study is not affected by these factors because these regions are automatically excluded by facial skin area segmentation and the reconstructed 3DMM.

Our scan dataset has limitations in the diversity of faces, leading to similarities in the reconstructed shapes and texture maps. For the BareSkinNet module, we believe the BFM~\cite{5279762} or FLAME~\cite{FLAME:SiggraphAsia2017} model can be employed, which is more suitable for non-Asian faces, instead of our original 3DMM. But when considering subsequent applications, we propose that 3DMM originates from the same scan dataset with high-resolution texture maps because consistency between 3DMM, bare skin image, and texture maps are preferred.

\section{Conclusions and Future Work}
We presented BareSkinNet, a framework to remove makeup and lighting influences from the face image input to reconstruct a high-fidelity 3D face model. Using our BareSkinNet as a preprocessing step for 3D face reconstruction, we can obtain consistent results of high-fidelity texture maps for the same subject. Through experiments, we confirmed that our approach could be successfully applied to various makeup image inputs.

\begin{figure}[t]
    \centering
    \includegraphics[width=\linewidth]{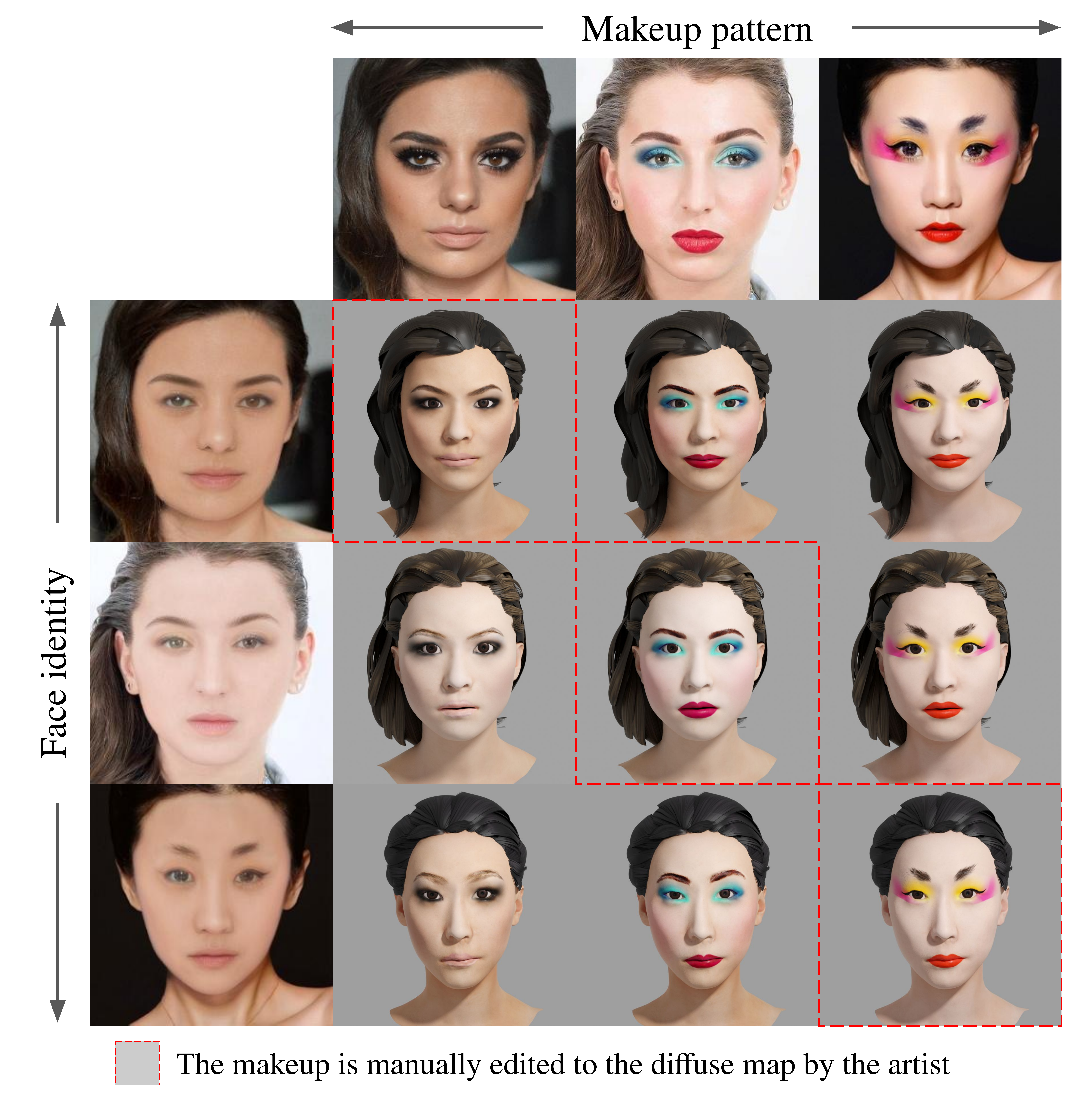}
    \caption{Examples of makeup transfer. Images with the red square show makeup results by editing diffuse texture maps by the 3DCG artist. Other images show the results of makeup transfer.} 
    \vspace{-3mm}
\label{fig:08_makeup_editing}
\end{figure}

Fig. \ref{fig:08_makeup_editing} shows an example of makeup editing. The reconstructed face models can be found in Figs. \ref{fig:teaser} and \ref{fig:04_result}. In this example, we asked a 3DCG artist to put makeup on the reconstructed face models by editing the inferred diffuse texture maps. Note that the 3DCG artist also created eyes and hairs. Created models are marked with the red square in the figure. A makeup layer was extracted by subtracting the reconstructed and edited diffuse texture. The makeup transfer was then achieved by adding the reconstructed diffuse texture maps and extracted makeup layers. The images without the red square show the results of makeup transfer. These results confirm the possibility of creating photo-realistic avatars in various makeup styles with minor effort.

This study focused on de-makeup and de-lighting for 3D face reconstruction. Currently, our system cannot separate makeup style and skin color. In the future, we plan to extend our method to extract the makeup layer and use it for 3D makeup reconstruction and transfer.

In addition, there is a potential to use BareSkinNet for other tasks such as face recognition and face verification. It can also be used to improve the accuracy of 3DMM texture space, for example, the texture creation process of the FLAME~\cite{FLAME:SiggraphAsia2017} model. We will investigate the availability of our method.
\vspace{-3mm}
\section*{Acknowledgements}
We thank Vladlen Erium for providing the 3D scan dataset, as well as Vladislava Mironenko for her help with 3D makeup editing. This work was supported by CyberHuman Productions.
\vspace{-3mm}

\bibliographystyle{eg-alpha-doi}  
\bibliography{main}        


\end{document}